%% file: main.tex
\documentclass[runningheads,table,xcdraw]{llncs}
\usepackage{pgfplots} 
\usepackage{graphicx}
\usepackage{subfiles}
\usepackage{hyperref}
\usepackage{tabularx}
\usepackage{placeins}
\usepackage{float}
\usepackage{svg}

\begin{document}
\title{Information Redundancy and Biases in Public Document Information Extraction Benchmarks}
\author{Seif Laatiri, Pirashanth Ratnamogan, Joël Tang, Laurent Lam, William Vanhuffel, Fabien Caspani}
\authorrunning{S. Laatiri et al.}
\titlerunning{Information redundancy in public KIE benchmarks}
\institute{BNP Paribas 
(seifedinne.laatiri;pirashanth.ratnamogan;joel.tang;laurent.lam)@bnpparibas.com
(william.vanhuffel;fabien.caspani)@bnpparibas.com}
\maketitle              
\begin{abstract}
\subfile{Sections/1-Abstract}
\end{abstract}
\section{Introduction}
\subfile{Sections/2-Introduction}

\section{Background}

\subsection{Related work}
\subfile{Sections/3-RelatedWork}

\subsection{Datasets}
\subfile{Sections/4-Datasets}

\subsection{Problem statement}
\subfile{Sections/5-ProblemStatement}

\section{Approach}
\subfile{Sections/6-Approach}

\section{Experiments}
\subfile{Sections/7-Experiments}

\FloatBarrier
\section{Conclusion}
\subfile{Sections/8-Conclusion}

\bibliography{main}
\bibliographystyle{splncs04}

\end{document}

%% file: Sections/1-Abstract.tex
Advances in the Visually-rich Document Understanding (VrDU) field and particularly the Key-Information Extraction (KIE) task are marked with the emergence of efficient Transformer-based approaches such as the LayoutLM models. Despite the good performance of KIE models when fine-tuned on public benchmarks, they still struggle to generalize on complex real-life use-cases lacking sufficient document annotations. Our research highlighted that KIE standard benchmarks such as SROIE and FUNSD contain significant similarity between training and testing documents and can be adjusted to better evaluate the generalization of models.\\
In this work, we designed experiments to quantify the information redundancy in public benchmarks, revealing a 75\% template replication in SROIE's official test set and 16\% in FUNSD's. We also proposed re-sampling strategies to provide benchmarks more representative of the generalization ability of models. We showed that models not suited for document analysis struggle on the adjusted splits dropping on average 10,5\% F1 score on SROIE and 3.5\% on FUNSD compared to multi-modal models dropping only 7,5\% F1 on SROIE and 0.5\% F1 on FUNSD.

\keywords{Visually-rich Document Understanding \and Key Information Extraction \and Named Entity Recognition \and Generalization Assessment.}

%% file: Sections/2-Introduction.tex
Visually-rich Document Understanding (VrDU) is a field that has seen progress lately following the recent breakthroughs in Natural Language Processing and Computer Vision. This field aims to transform documents into structured data by simultaneously leveraging their textual, positional and visual attributes. Since scanned documents are often noisy, recent works addressed VrDU as a two component stream: first extracting the text with optical character recognition and then performing analysis using OCR text, document’s layout and visual attributes.

Real life business documents can belong to multiple categories such as financial reports, employee contracts, forms, emails, letters, receipts, resumes and others. Thus it is challenging to create general pipelines capable of handling all types of documents. State-of-the-art models focus on document-level pre-training objectives then fine-tuning on downstream tasks. These models show promising results on public information extraction benchmarks such as FUNSD \cite{funsd}, SROIE \cite{sroie}, CORD \cite{cord}, Kleister NDA \cite{Kleister}. However, on real world complex use cases it is difficult to replicate the same performances due to a lack of annotated samples and diversity in their templates. In this paper, we show that common benchmarks can be managed to better evaluate the generalization ability of information extraction models and thus become more viable tools for model selection for real-world business use cases.

To this end, we focus on SROIE \cite{sroie} and FUNSD \cite{funsd} and explore their potential to challenge the generalization power of a given model. We design experiments to measure document similarities in the official training and testing splits of these benchmarks and propose resampling strategies to render these benchmarks a better evaluation of models’ performance on unseen documents. We then investigate the impact of these resampling strategies on state-of-the-art VrDU models.

%% file: Sections/3-RelatedWork.tex
\subsubsection{Dataset biases and model generalization in NLP} 

The study of dataset biases and model generalization is an important area of research that has already been conducted on several NLP tasks. In the Named Entity Recognition task, several studies have highlighted the fact that the common datasets CONLL 2003 \cite{sang2003introduction} and OntoNotes 5 \cite{weischedel2013ontonotes} are strongly biased by an unrealistic lexical overlap between mentions in training and test sets \cite{augenstein2017generalisation,taille2020contextualized}.
In the co-reference task, the same lexical overlap with respect to co-reference mentions was observed in CONLL 2012 \cite{pradhan2012conll} and led to an overestimation of the performance of deep learning models compared to classical methods for real world applications \cite{moosavi2017using}. The Natural Language Inference (NLI) task also suffers from a dataset with bias. Indeed, MultiNLI has been reported to suffer from both lexical overlap and hypothesis-only bias: the fact that hypothesis sentences contain words associated with a target label \cite{gururangan-etal-2018-annotation}.

 Deep Learning models are extremely sensitive to these biases. However in real world, models should be able to generalize to new out of domain data. Multiple studies experimented models memorization capability \cite{arpit2017closer} and models robustness in order to perform on out of domain data in multiple tasks: translation \cite{DBLP:conf/ecai/MghabbarR20}, co-reference \cite{toshniwal-etal-2021-generalization} or named entity recognition \cite{taille2020contextualized}.\\
To the best of our knowledge, our work is the first one assessing and analyzing biases in the context of information extraction from documents. 

\subsubsection{Information extraction models}

When performing visual analysis on documents, early work handled separately the different modalities. Preliminary work \cite{Hao,Schreiber} focused on extracting tabular data from documents by combining heuristic
rules and convolutional networks to detect and recognize tabular data. Later work \cite{soto} used visual features for document layout detection by incorporating contextual information in Faster R-CNNs \cite{Ren}. Follow up research \cite{Liu,yu2020} combined textual and visual information by introducing a graph convolution based model for key information extraction. This approach exhibited good results however, models are only using supervised data which is limited. In addition, these pre-training methods do not inherently combine textual and visual attributes as they are merged during fine-tuning instead.

Following the rise of Transformers, more Transformer based models were adapted for VrDU with novel pre-training objectives. LayoutLM \cite{layoutlm} uses 2D positional embeddings in order to integrate layout information with word embeddings and is pretrained on layout understanding tasks. LayoutLMv2 \cite{layoutlmv2} adds token-level visual features extracted with Convolution Neural Networks and models interactions among text, layout and image. Later work aimed to match the reconstruction pre-training objectives of masked text with a similar objective for reconstructing visual attributes such as LayoutLMv3 \cite{layoutlmv3} which proposed to predict masked image areas through a set of image tokens similarly to visual Transformers \cite{vit,vilt}. Other recent approaches \cite{donut,pixstruct} experimented with an OCR-free setup by leveraging an encoder-decoder visual transformer.

%% file: Sections/4-Datasets.tex
Multiple datasets exist to benchmark a variety of document understanding tasks. For instance RVL-CDIP \cite{rvlcdip} is an image-centric dataset for document image classification, FUNSD \cite{funsd}, CORD \cite{cord}, SROIE \cite{sroie} and Kleister-NDA \cite{Kleister} are datasets for key information extraction respectively from forms, receipts and contracts whilst DocVQA \cite{docvqa} is a benchmark for visual question answering.

In this work, we focus on the task of information extraction while investigating template similarities in the current documents distribution within the datasets. Documents having the same template are documents sharing the same layout that can be read in the same way. We decided to primarily work with SROIE and FUNSD as they are common benchmarks displaying two different types of documents. In more details, the SROIE dataset for Scanned Receipt OCR and Information Extraction was presented in the 2019 edition of the ICDAR conference. It represents processes of recognizing text from scanned restaurant receipts and extracting key entities from them. The dataset contains 1000 annotated scanned restaurant receipts split into train/test splits with the ratio 650/350. Three tasks were set up for the competition: Scanned Receipt Text Localisation, Scanned Receipt Optical Character Recognition and Key Information Extraction from Scanned Receipts.

Second, FUNSD is a dataset for Form Understanding in Noisy Scanned Documents that have been a constant benchmark in recent document information extraction work. It contains noisy scanned forms and aims at extracting and structuring their textual contents. The dataset provides annotations for text recognition, entity linking and semantic entity labeling. In this work, FUNSD refers to the revised version of the dataset released by the authors \cite{revisedFunsd} containing the same documents with cleaned annotations.

%% file: Sections/5-ProblemStatement.tex
We formalize key information extraction as a token classification task on the tokenized text extracted from the document.\\   
Let us denote by $T = t_{0<i\leq n}$ the sequence of text tokens $t_i$ extracted from a document $\mathcal{D}$. Let $\mathcal{I}$ be the image of the document $\mathcal{D}$ and $m$ the number of entity types (since we perform Inside-Outside-Beginning tagging \cite{iobtagging} similarly to named entity recognition tasks, the number of entity classes is $2m+1$). We aim to build a classifier $\mathcal{F}$ such that for every token $t_i$ in the sequence:  
\begin{equation} 
\mathcal{F}(t_i | T,\mathcal{I}) = c 
\end{equation} 
with $c \in \{1, \ldots, 2m+1 \}$ the target IOB class of token $t_i$.

In IOB tagging, an entity spans over multiple adjacent tokens $(t_i)_{i_{start}\leq i\leq i_{end}}$ and is only correctly predicted if all of its tokens are correctly predicted, that is \begin{math} \mathcal{F}(t_{i_{start}} | T,\mathcal{I}) = B-entity \quad and \quad \mathcal{F}(t_i | T,\mathcal{I}) = I-entity  \; for  \; i \in  \{i_{start}+1, \ldots, i_{end}\}\end{math}

%% file: Sections/6-Approach.tex
\subsection{Motivation}
Recent document analysis models such as LayoutLM models \cite{layoutlm,layoutlmv2,layoutlmv3}, DocFormer \cite{docformer}, Lilt \cite{lilt} and others used the datasets mentioned above to evaluate their models and benchmark them against other state-of-the-art works. However, these evaluation metrics are usually difficult to replicate on real-life business use-cases, particularly those obtained on SROIE and FUNSD. This discrepancy is, in part, due to the complexity of business use-cases and their lack of good-quality annotated data. However we also suspect that the current train and test splitting distribution of these datasets does not optimally evaluate the ability of models to generalize on unseen documents. 

Even though it is a common practice in machine learning to keep similar distributions for both training and testing data, this practice is not optimal when benchmarking and comparing models that will later be finetuned on small datasets or inferred on out-of-domain data. By containing similar documents in both training and testing data, these datasets allow models to memorize predictions during training and simply infer them on test documents without evaluating their ability to understand and analyze new templates of documents. In real business use-cases, this is particularly harming long-term performances, as domain shift often occurs after a certain period of time, when new unseen templates are used.

\subsection{Resampling datasets}
For each studied dataset, the current training and testing documents are thoroughly observed and analysed for homogeneous samples. We customize for every dataset a method to group similar documents and re-sample the training and testing splits to minimize template similarity and redundancy. We remind that the term template in this context refers to the disposition and layout of a document. 

\subsubsection{SROIE:}
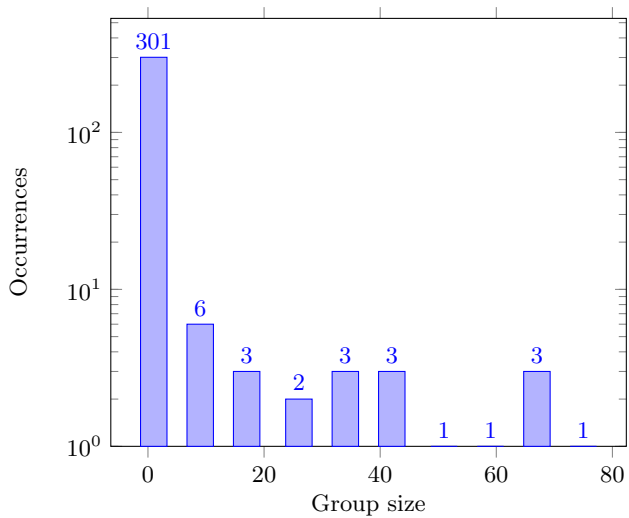
\begin{figure}[htbp]
\centering
    \begin{tikzpicture}
        \begin{semilogyaxis}[ybar, ymin=1, xlabel=Group size, ylabel=Occurrences,nodes near coords, point meta=rawy]
            \addplot coordinates { (1,301) (9,6) (17,3) (26,2) (34,3) (42,3) (51,1) (59,1) (67,3) (75,1)};
        \end{semilogyaxis}
    \end{tikzpicture}
    \caption{SROIE: distribution of similarity groups, each group containing receipts of the same template}    
    \label{fig:sroie_groups_distribution}
\end{figure}

\begin{figure}[htbp]
` \centering
  \includegraphics[width = 12cm]{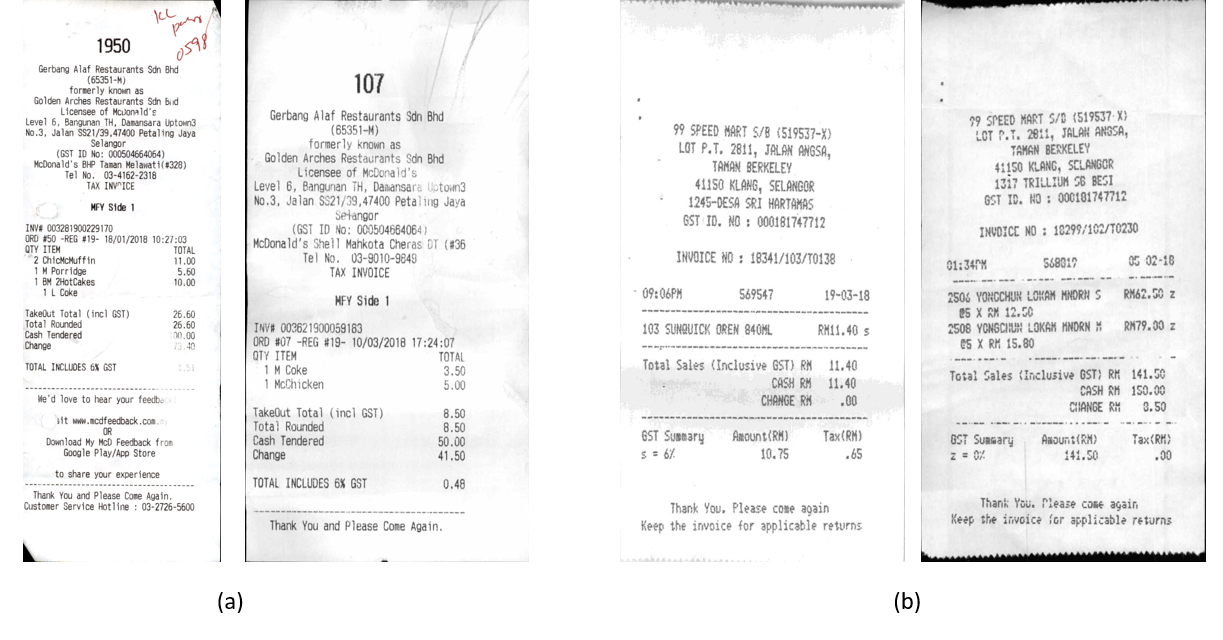}
  \caption{SROIE: Receipts from two different business (a) and (b)}
  \label{fig:sroie_samples}
\end{figure}

Information extraction in this dataset is performed by extracting semantic entities from business receipts such as business' name, address, the order's date and total price. As shown in Figure \ref{fig:sroie_samples}, receipts from the same business have a similar disposition, they contain the same business' name and address as well as the same template. The current official data split of SROIE does not account for this factor as same business receipts can be present in both train and test documents. We group same businesses and re-sample train-test splits while assuring that every group is present in only a single split. The sizes of groups of samples sharing the same template varies from 1 to 76 receipts and their distribution is described in Figure \ref{fig:sroie_groups_distribution} in a logarithmic scale.  

\subsubsection{FUNSD: }
\begin{figure}[htbp]
` \centering
  \includegraphics[width = 14cm]{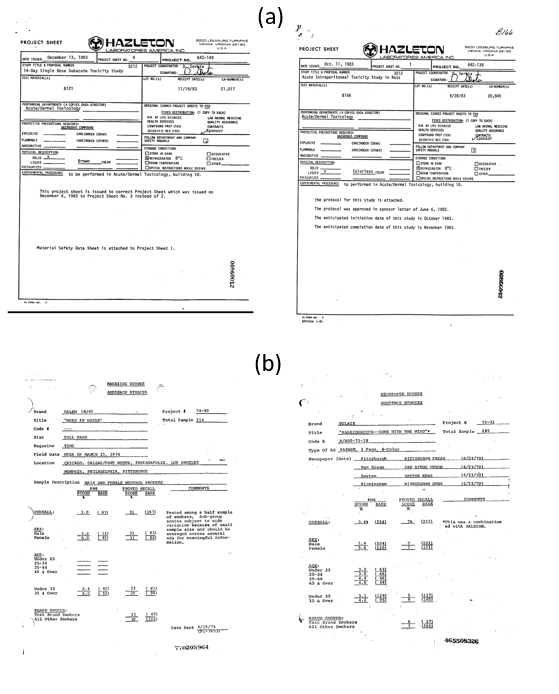}
  \caption{FUNSD: Forms from two different templates (a) and (b)}
  \label{fig:funsd_samples}
\end{figure}

By investigating FUNSD, we observed that multiple forms of the same template were present in the dataset while being filled with different information, for instance a standard hospital application filled by different patients. Having the exact same form template in both training and testing is a clear indicator of information redundancy as illustrated in the template comparison in Figure \ref{fig:funsd_samples}. Based on the fact that same template forms share similar slot names (\textit{questions}), we introduced a similarity metric on forms using the overlap of their question annotations. Based on this assumption, we propose the following overlap score for two forms:
\[Overlap(docA,docB) = \frac{Count(QuestionsA \cap QuestionsB)} {Max(len(QuestionsA), len(QuestionsB))}\]
where $QuestionsA$ and $QuestionsB$ are respectively the question annotations sets of document A and document B.

We have manually defined a set of template groups as ground truths and then evaluated different grouping similarity thresholds, eventually keeping a threshhold of 0.7. This metric was next used to group forms of same templates. The sizes of groups in this case was far lower than that of SROIE groups as the biggest group of forms sharing the same template was limited to 4 forms. From 50 forms in the testing set, we found 8 (16\%) that shared the same template with at least one training form. We resample the train and test splits accordingly, ensuring that no forms with the same template are present in both splits.
The resampled splits can be found in \href{https://github.com/Seif-Lat/Bias-study-FUNSD-SROIE}{https://github.com/Seif-Lat}
\begin{figure}[htbp]
\centering
    \begin{tikzpicture}
        \begin{axis}[ybar, ymin=1, bar width=25pt , xtick={1,2,3,4}, xlabel=Group size, ylabel=Occurrences, nodes near coords]
            \addplot coordinates { (1, 130) (2, 21) (3, 8) (4,1)};
        \end{axis}
    \end{tikzpicture}
    \caption{FUNSD: distribution of similarity groups, each group containing forms of the same template}    
    \label{fig:funsd_groups_distribution}
\end{figure}
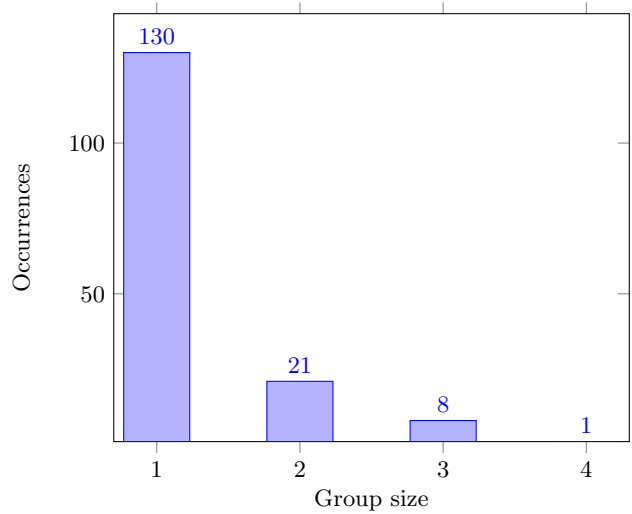

\subsection{Models}

In the context of information extraction from documents seen as a token classification task, three approaches exist in the literature:
\begin{itemize}
    \item Standard NLP models using textual information,
    \item Layout Aware models using both text and layout information,
    \item Multi-modal approaches using text, layout and visual representations of tokens.
\end{itemize}

\begin{figure}[htbp]
` \centering
  \includegraphics[width = 14cm]{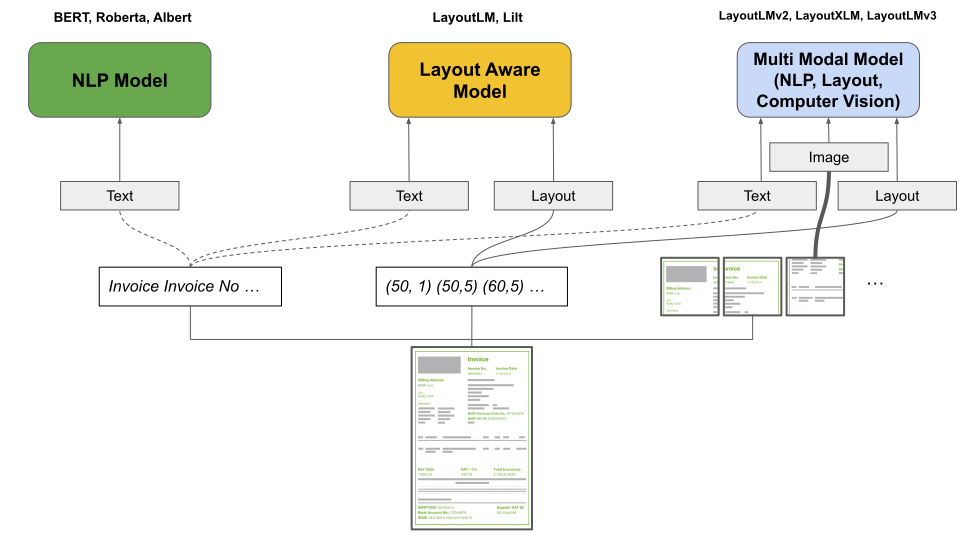}
  \caption{Overview of document understanding models}
\end{figure}

In the context of our study, it is important to challenge the common evidence obtained using official datasets: multi-modal approaches are more effective than other approaches.

We studied the fine-tuning of the following pretrained models:

\subsubsection{BERT} \cite{bert} is a bidirectional Transformer-based language model pretrained on a large corpus with Mask Language Modeling (MLM) and Next Sentence Prediction (NSP) tasks.
\subsubsection{RoBERTa} \cite{roberta} differs from BERT during pretraining with the use of a significantly larger dataset, a larger batch size and dynamic masking in the MLM task while dropping the NSP pretraining objective.

\subsubsection{AlBERT} \cite{lan2019albert} is a scalable version of BERT that uses two methods to reduce the model memory footprint: sharing some of the model layers, and a factorized embedding parameterization.

\subsubsection{Lilt} \cite{wang2022lilt} is an approach decoupling layout and text representation in order to have the model using more layout information and being more language independent. It uses independent layout and text pretraining but also proposes a bi-direction attention complementation mechanism in order to combine two flows: one for layout the other one for the text.
\subsubsection{LayoutLM} \cite{layoutlm} introduced a multimodal pretraining approach combining text and layout features for document image understanding and information extraction tasks. It leverages both text and layout features and incorporates them into a single framework.
\subsubsection{LayoutLMv2} \cite{layoutlmv2}
LayoutLMv2 is one of the first Transformers to use image feature during the pretraining process. They propose to use the output of a CNN architecture to create image token embeddings which gives useful information about the document layout. 
\subsubsection{LayoutXLM} \cite{layoutxlm}
LayoutXLM has the same architecture as LayoutLMv2 but is pretrained on a multi-lingual dataset. 
\subsubsection{LayoutLMv3} \cite{layoutlmv3}
LayoutLMv3 uses a multi modal Transformer architecture and introduces an image reconstruction pretraining objective similar to text reconstruction in the masked language modeling objective.

%% file: Sections/7-Experiments.tex
\subsection{Experimental setup}
We use the same configuration when fine-tuning all the models. We use a batch size of 2 and an Adam optimizer with an initial learning rate of $2*10^{-5}$.
We decrease the learning rate by half every 10 epochs without improvement in the validation F1 score. We stop the fine-tuning when the learning rate goes below $10^{-7}$. We finally recover the model with the best validation F1.

For each experiment, the test set is defined initially, either the official testing set on the original datasets or our extracted testing set on the resampled datasets. We then generate four different splits from the remaining data as training and validation data with 80-20 ratio, we train and test each model on the four splits and present the average performance in the sections below. We perform this cross validation for a more robust model comparison as we have observed a shift of performance in consecutive trainings of the same model.

\subsection{Results on original datasets}
We train a group of models leveraging different modalities on receipt understanding and form understanding using the official splits of SROIE and FUNSD. Results are presented in table \ref{tab:original_split}. On form understanding (FUNSD), models using only the textual information of documents (BERT, AlBERT, RoBERTa) perform marginally worse than multi-modal models as their F1 scores are on average 20 points less. RoBERTa has the highest F1 (80.64) score among textual models and is 5 points behind the closest multi-modal model being LayoutXLM with 85.57 F1. Among multi-modal models, LayoutLMv2 and LayoutLMv3 have better scores than LayoutLM and LiLT since they also leverage visual attributes of forms. LayoutLMv3 has the highest F1 score of 88.81 thanks to its efficient multi-modal pre-training. \\

On receipt understanding (SROIE) multi-modal approaches also have higher metrics than textual models, however the discrepancy between them is much lower as the F1 increase between each multi-modal and the average score of all textual models varies from 2 to 4 and is considerably lower than what we observed on FUNSD. On this task LayoutLMv2 achieves the highest score amongst all models, reaching 96.14 F1. These results validate the importance of positional and visual attributes in information extraction tasks on visually rich documents.   

\begin{table}[!htbp]
\resizebox{\columnwidth}{!}{%
\begin{tabular}{c|c|ccc|ccc}
\hline
              & & \multicolumn{3}{c}{\textbf{SROIE}} & \multicolumn{3}{c}{\textbf{FUNSD}} \\ \hline
              \textbf{Model} & \textbf{Params} & \textbf{F1}  & \textbf{Precision} & \textbf{Recall}
              & \textbf{F1}   & \textbf{Precision} & \textbf{Recall}\\ 
      \hline
      BERT$_{base}$ & 110M & 92.47 & 92.36 & 92.68 & 61.03 & 60.29 & 61.90 \\
      AlBERT$_{base}$ & 12M & 92.28 & 92.28 & 92.28 & 57.39 & 56.08 & 58.99 \\
      RoBERTa$_{base}$ & 125M & 93.90 & 93.23 & 94.61 & 80.64 & 81.36 & 79.96 \\
      LayoutLM$_{base}$ & 112M & 94.57 & 93.93 & 95.22 & 86.07 & 86.24 & 85.63 \\
      LiLT$_{base}$ & 131M & 95.41 & 95.23 & 95.60 & 87.41 & 87.41 & 87.41\\
      LayoutLMv2$_{base}$ & 200M & \textbf{96.14} & \textbf{96.39} & \textbf{95.94} & 88.14 & 88.31 & 88.08\\
      LayoutXLM$_{base}$ & 369M & 94.75 & 94.57 & 94.94 & 85.57 & 85.87 & 85.46\\
      LayoutLMv3$_{base}$ & 126M & 95.11 & 94.87 & 95.70 & \textbf{88.81} & \textbf{89.32} & \textbf{88.46}\\
      \hline
      BERT$_{large}$ & 351M & 93.72 & 93.47 & 94.02 & 61.03 & 60.29 & 61.90 \\
      AlBERT$_{large}$ & 18M & 88.96 & 87.86 & 90.20 & 58.81 & 57.46 & 60.39 \\
      RoBERTa$_{large}$ & 355M & 94.99 & 94.91 & 95.09 & 82.43 & 82.99 & 81.78 \\
      LayoutLM$_{large}$ & 340M & 94.70 & 94.31 & 95.10 & 84.29 & 84.60 & 84.10\\
      LayoutLMv2$_{large}$ & 426M & \textbf{96.55} & \textbf{96.69} & \textbf{96.42} &  88.79 & 89.06 & 88.75 \\
      LayoutLMv3$_{large}$ & 365M & 95.87 & 95.71 & 96.03 & \textbf{89.84} & \textbf{89.97} & \textbf{89.56}\\
\end{tabular}%
}
\caption{Performance of state-of-the-art information extraction models on SROIE and FUNSD official testing sets.} 
\label{tab:original_split}
\end{table}

\subsection{Results on resampled datasets}
After having resampled the splits of both FUNSD and SROIE, we train the same group of models and present results in table \ref{tab:domain_ad}. On form understanding, textual models are outperformed by multi-modal models and show a more important decrease in performance compared to results on the original split with an average drop of 3.5 F1 score compared to only 0.5 in multi-modal models.

\begin{table}[!htbp]
\resizebox{\columnwidth}{!}{%
\begin{tabular}{c|c|ccc|ccc}
\hline
              & & \multicolumn{3}{c}{\textbf{SROIE}} & \multicolumn{3}{c}{\textbf{FUNSD}} \\ \hline
              \textbf{Model} & \textbf{Params} & \textbf{F1}   & \textbf{Precision} & \textbf{Recall}         &\textbf{F1}   & \textbf{Precision} & \textbf{Recall} \\
      \hline
      BERT$_{base}$ & 110M & 81.00 & 77.04 & 86.21 & 55.25 & 54.97 & 55.86\\
      AlBERT$_{base}$ & 12M & 79.86 & 77.48 & 82.97 & 53.66 & 52.63 & 54.99 \\
      RoBERTa$_{base}$ & 125M & 86.05 & 84.15 & 88.20 & 78.71 & 79.16 & 78.59 \\
      LayoutLM$_{base}$ & 112M & 84.80 & 82.18 & 88.02 & 85.88 & 85.85 & 86.11 \\
      LiLT$_{base}$ & 131M & \textbf{89.38} & 86.94 & \textbf{92.11} & 84.76 & 85.12 & 84.91 \\
      LayoutLMv2$_{base}$ & 200M & 87.92 & 86.67 & 89.34 & 88.61 & 88.65 & 88.94\\
      LayoutXLM$_{base}$ & 369M & 87.99 & 86.24 & 90.14 & 85.51 & 85.98 & 86.25 \\
      LayoutLMv3$_{base}$ & 126M & 87.86 & \textbf{87.29} & 88.71 & \textbf{89.07} & \textbf{89.26} & \textbf{89.16}\\
      \hline
      BERT$_{large}$ & 351M & 81.15 & 77.53 & 85.84 & 55.90 & 55.46 & 56.59\\
      AlBERT$_{large}$ & 18M & 82.82 & 81.16 & 84.76 & 54.93 & 52.73 & 57.77 \\
      RoBERTa$_{large}$ & 355M & 87.86 & 86.16 & 89.84 & 80.28 & 80.16 & 80.66\\
      LayoutLM$_{large}$ & 340M & 84.78 & 82.62 & 87.91 & 85.95 & 85.75 & 86.34\\
      LayoutLMv2$_{large}$ & 426M & 87.98 & 86.71 & 89.98 & \textbf{90.14} & \textbf{90.19} & \textbf{90.31}\\
      LayoutLMv3$_{large}$ & 356M & \textbf{88.47} & \textbf{87.00} & \textbf{90.13} & 89.86 & 89.72 & 90.31\\
\end{tabular}%
}
\caption{Performance of state-of-the-art information extraction models on SROIE and FUNSD \textbf{resampled} testing sets.}
\label{tab:domain_ad}
\end{table}

On receipt understanding, F1 scores drastically drop compared to results on the original split in table \ref{tab:original_split}, BERT, AlBERT and RoBERTa drop on average 10.5 F1 points whereas multi-modal models drop only 7.5 F1 points on average. Scores on the adjusted splits show a higher discrepancy between the models and are more consistent with the modalities leveraged by each model and the efficiency of their pretraining. For instance LiLT marginally outperforms LayoutLM as its pretraining allows it to better leverage the positional information of documents. The adjusted split also makes the information extraction task more challenging as the average F1 score drops from 94.34 to 85.60 and the highest reached F1 drops from 96.55 to 89.38.

These results show that the original splits of both SROIE and FUNSD contained data leaks that allowed models to infer on testing data without necessarily learning how to understand new templates. The adjusted splits evaluate more properly the generalization ability of models and their capacity to transfer knowledge to unseen templates.

%% file: Sections/8-Conclusion.tex
In this paper, we showed that SROIE and FUNSD were featuring information redundancies between their training and testing sets. Having trained multiple state-of-the-art models on VrDU tasks, we showed that this information redundancy artificially increases the models performances. In particular, we observe that SROIE is still a challenging benchmark as the average F1 score drops from 96.38 on the official splits to 88.78 on the adjusted splits, proving that it remains a viable benchmark for upcoming works when it is sampled more carefully. 

These findings demonstrate that generating Independent and Identically Distributed splits for evaluation datasets as is traditionally done is not an optimal approach, as it introduces a high memorization bias especially with large neural networks. The 0\% overlap approach presented in this work is an example of an alternate strategy specific to evaluating models’ generalization on unseen templates and is closer to real-world use-cases than traditional splits. Other criteria can also be explored for this same purpose such as a date based resampling.

%% file: main.bbl
\begin{thebibliography}{10}
\providecommand{\url}[1]{\texttt{#1}}
\providecommand{\urlprefix}{URL }
\providecommand{\doi}[1]{https://doi.org/#1}

\bibitem{docformer}
Appalaraju, S., Jasani, B., Kota, B.U., Xie, Y., Manmatha, R.: Docformer:
  End-to-end transformer for document understanding. CoRR
  \textbf{abs/2106.11539} (2021), \url{https://arxiv.org/abs/2106.11539}

\bibitem{arpit2017closer}
Arpit, D., Jastrzebski, S., Ballas, N., Krueger, D., Bengio, E., Kanwal, M.S.,
  Maharaj, T., Fischer, A., Courville, A., Bengio, Y., et~al.: A closer look at
  memorization in deep networks. In: International conference on machine
  learning. pp. 233--242. PMLR (2017)

\bibitem{augenstein2017generalisation}
Augenstein, I., Derczynski, L., Bontcheva, K.: Generalisation in named entity
  recognition: A quantitative analysis. Computer Speech \& Language
  \textbf{44},  61--83 (2017)

\bibitem{bert}
Devlin, J., Chang, M.W., Lee, K., Toutanova, K.: Bert: Pre-training of deep
  bidirectional transformers for language understanding  \textbf{Proceedings of
  the 2019 Conference of the North American Chapter of the Association for
  Computational Linguistics: Human Language Technologies, Volume 1 (Long and
  Short Papers)} (2019), \url{https://aclanthology.org/N19-1423.pdf}

\bibitem{vit}
Dosovitskiy, A., Beyer, L., Kolesnikov, A., Weissenborn, D., Zhai, X.,
  Unterthiner, T., Dehghani, M., Minderer, M., Heigold, G., Gelly, S.,
  Uszkoreit, J., Houlsby, N.: An image is worth 16x16 words: Transformers for
  image recognition at scale  (2020), \url{https://arxiv.org/abs/2010.11929}

\bibitem{gururangan-etal-2018-annotation}
Gururangan, S., Swayamdipta, S., Levy, O., Schwartz, R., Bowman, S., Smith,
  N.A.: Annotation artifacts in natural language inference data. In:
  Proceedings of the 2018 Conference of the North {A}merican Chapter of the
  Association for Computational Linguistics: Human Language Technologies,
  Volume 2 (Short Papers). pp. 107--112. Association for Computational
  Linguistics, New Orleans, Louisiana (Jun 2018). \doi{10.18653/v1/N18-2017},
  \url{https://aclanthology.org/N18-2017}

\bibitem{Hao}
Hao, L., Gao, L., Yi, X., Tang, Z.: A table detection method for pdf documents
  based on convolutional neural networks. In: 2016 12th IAPR Workshop on
  Document Analysis Systems (DAS). pp. 287--292 (2016).
  \doi{10.1109/DAS.2016.23}

\bibitem{rvlcdip}
Harley, A.W., Ufkes, A., Derpanis, K.G.: Evaluation of deep convolutional nets
  for document image classification and retrieval. CoRR
  \textbf{abs/1502.07058} (2015), \url{http://arxiv.org/abs/1502.07058}

\bibitem{layoutlmv3}
Huang, Y., Lv, T., Cui, L., Lu, Y., Wei, F.: Layoutlmv3: Pre-training for
  document ai with unified text and image masking. In: Proceedings of the 30th
  ACM International Conference on Multimedia. p. 4083–4091. MM '22,
  Association for Computing Machinery, New York, NY, USA (2022).
  \doi{10.1145/3503161.3548112}, \url{https://doi.org/10.1145/3503161.3548112}

\bibitem{sroie}
Huang, Z., Chen, K., He, J., Bai, X., Karatzas, D., Lu, S., Jawahar, C.:
  Icdar2019 competition on scanned receipt ocr and information extraction pp.
  1516--1520 (2019), \url{https://arxiv.org/pdf/2103.10213.pdf}

\bibitem{funsd}
Jaume, G., Kemal~Ekenel, H., Thiran, J.P.: Funsd: A dataset for form
  understanding in noisy scanned documents. In: 2019 International Conference
  on Document Analysis and Recognition Workshops (ICDARW). vol.~2, pp.~1--6
  (2019). \doi{10.1109/ICDARW.2019.10029}

\bibitem{donut}
Kim, G., Hong, T., Yim, M., Nam, J., Park, J., Yim, J., Hwang, W., Yun, S.,
  Han, D., Park, S.: Ocr-free document understanding transformer (2022)

\bibitem{vilt}
Kim, W., Son, B., Kim, I.: Vilt: Vision-and-language transformer without
  convolution or region supervision (2021),
  \url{https://arxiv.org/abs/2102.03334}

\bibitem{lan2019albert}
Lan, Z., Chen, M., Goodman, S., Gimpel, K., Sharma, P., Soricut, R.: Albert: A
  lite bert for self-supervised learning of language representations. arXiv
  preprint arXiv:1909.11942  (2019)

\bibitem{pixstruct}
Lee, K., Joshi, M., Turc, I., Hu, H., Liu, F., Eisenschlos, J., Khandelwal, U.,
  Shaw, P., Chang, M.W., Toutanova, K.: Pix2struct: Screenshot parsing as
  pretraining for visual language understanding (2022)

\bibitem{Liu}
Liu, X., Gao, F., Zhang, Q., Zhao, H.: Graph convolution for multimodal
  information extraction from visually rich documents. In: NAACL (2019)

\bibitem{roberta}
Liu, Y., Ott, M., Goyal, N., Du, J., Joshi, M., Chen, D., Levy, O., Lewis, M.,
  Zettlemoyer, L., Stoyanov, V.: Roberta: A robustly optimized bert pretraining
  approach (2019). \doi{10.48550/ARXIV.1907.11692},
  \url{https://arxiv.org/abs/1907.11692}

\bibitem{docvqa}
Mathew, M., Karatzas, D., Manmatha, R., Jawahar, C.V.: Docvqa: {A} dataset for
  {VQA} on document images. CoRR  \textbf{abs/2007.00398} (2020),
  \url{https://arxiv.org/abs/2007.00398}

\bibitem{DBLP:conf/ecai/MghabbarR20}
Mghabbar, I., Ratnamogan, P.: Building a multi-domain neural machine
  translation model using knowledge distillation. In: Giacomo, G.D.,
  Catal{\'{a}}, A., Dilkina, B., Milano, M., Barro, S., Bugar{\'{\i}}n, A.,
  Lang, J. (eds.) {ECAI} 2020 - 24th European Conference on Artificial
  Intelligence, 29 August-8 September 2020, Santiago de Compostela, Spain,
  August 29 - September 8, 2020 - Including 10th Conference on Prestigious
  Applications of Artificial Intelligence {(PAIS} 2020). Frontiers in
  Artificial Intelligence and Applications, vol.~325, pp. 2116--2123. {IOS}
  Press (2020). \doi{10.3233/FAIA200335},
  \url{https://doi.org/10.3233/FAIA200335}

\bibitem{moosavi2017using}
Moosavi, N.S., Strube, M.: Using linguistic features to improve the
  generalization capability of neural coreference resolvers. arXiv preprint
  arXiv:1708.00160  (2017)

\bibitem{cord}
Park, S., Shin, S., Lee, B., Lee, J., Surh, J., Seo, M., Lee, H.: Cord: A
  consolidated receipt dataset for post-ocr parsing  (2019)

\bibitem{pradhan2012conll}
Pradhan, S., Moschitti, A., Xue, N., Uryupina, O., Zhang, Y.: Conll-2012 shared
  task: Modeling multilingual unrestricted coreference in ontonotes. In: Joint
  Conference on EMNLP and CoNLL-Shared Task. pp. 1--40 (2012)

\bibitem{iobtagging}
Ramshaw, L., Marcus, M.: Text chunking using transformation-based learning. In:
  Third Workshop on Very Large Corpora (1995),
  \url{https://aclanthology.org/W95-0107}

\bibitem{Ren}
Ren, S., He, K., Girshick, R., Sun, J.: Faster r-cnn: Towards real-time object
  detection with region proposal networks. In: Cortes, C., Lawrence, N., Lee,
  D., Sugiyama, M., Garnett, R. (eds.) Advances in Neural Information
  Processing Systems. vol.~28. Curran Associates, Inc. (2015),
  \url{https://proceedings.neurips.cc/paper/2015/file/14bfa6bb14875e45bba028a21ed38046-Paper.pdf}

\bibitem{sang2003introduction}
Sang, E.F., De~Meulder, F.: Introduction to the conll-2003 shared task:
  Language-independent named entity recognition. arXiv preprint cs/0306050
  (2003)

\bibitem{Schreiber}
Schreiber, S., Agne, S., Wolf, I., Dengel, A., Ahmed, S.: Deepdesrt: Deep
  learning for detection and structure recognition of tables in document
  images. In: 2017 14th IAPR International Conference on Document Analysis and
  Recognition (ICDAR). vol.~01, pp. 1162--1167 (2017).
  \doi{10.1109/ICDAR.2017.192}

\bibitem{soto}
Soto, C., Yoo, S.: Visual detection with context for document layout analysis.
  In: Proceedings of the 2019 Conference on Empirical Methods in Natural
  Language Processing and the 9th International Joint Conference on Natural
  Language Processing (EMNLP-IJCNLP). pp. 3464--3470. Association for
  Computational Linguistics, Hong Kong, China (Nov 2019).
  \doi{10.18653/v1/D19-1348}, \url{https://aclanthology.org/D19-1348}

\bibitem{Kleister}
Stanislawek, T., Gralinski, F., Wr{\'{o}}blewska, A., Lipinski, D., Kaliska,
  A., Rosalska, P., Topolski, B., Biecek, P.: Kleister: Key information
  extraction datasets involving long documents with complex layouts. CoRR
  \textbf{abs/2105.05796} (2021), \url{https://arxiv.org/abs/2105.05796}

\bibitem{taille2020contextualized}
Taill{\'e}, B., Guigue, V., Gallinari, P.: Contextualized embeddings in
  named-entity recognition: An empirical study on generalization. In: European
  Conference on Information Retrieval. pp. 383--391. Springer (2020)

\bibitem{toshniwal-etal-2021-generalization}
Toshniwal, S., Xia, P., Wiseman, S., Livescu, K., Gimpel, K.: On generalization
  in coreference resolution. In: Proceedings of the Fourth Workshop on
  Computational Models of Reference, Anaphora and Coreference. pp. 111--120.
  Association for Computational Linguistics, Punta Cana, Dominican Republic
  (Nov 2021). \doi{10.18653/v1/2021.crac-1.12},
  \url{https://aclanthology.org/2021.crac-1.12}

\bibitem{revisedFunsd}
Vu, H.M., Nguyen, D.T.: Revising {FUNSD} dataset for key-value detection in
  document images. CoRR  \textbf{abs/2010.05322} (2020),
  \url{https://arxiv.org/abs/2010.05322}

\bibitem{lilt}
Wang, J., Jin, L., Ding, K.: Lilt: A simple yet effective language-independent
  layout transformer for structured document understanding (2022).
  \doi{10.48550/ARXIV.2202.13669}, \url{https://arxiv.org/abs/2202.13669}

\bibitem{wang2022lilt}
Wang, J., Jin, L., Ding, K.: Lilt: A simple yet effective language-independent
  layout transformer for structured document understanding. arXiv preprint
  arXiv:2202.13669  (2022)

\bibitem{weischedel2013ontonotes}
Weischedel, R., Palmer, M., Marcus, M., Hovy, E., Pradhan, S., Ramshaw, L.,
  Xue, N., Taylor, A., Kaufman, J., Franchini, M., et~al.: Ontonotes release
  5.0 ldc2013t19. Linguistic Data Consortium, Philadelphia, PA  \textbf{23}
  (2013)

\bibitem{layoutlmv2}
Xu, Y., Xu, Y., Lv, T., Cui, L., Wei, F., Wang, G., Lu, Y., Florencio, D.,
  Zhang, C., Che, W., Zhang, M., Zhou, L.: {L}ayout{LM}v2: Multi-modal
  pre-training for visually-rich document understanding. In: Proceedings of the
  59th Annual Meeting of the Association for Computational Linguistics and the
  11th International Joint Conference on Natural Language Processing (Volume 1:
  Long Papers). pp. 2579--2591. Association for Computational Linguistics,
  Online (Aug 2021). \doi{10.18653/v1/2021.acl-long.201},
  \url{https://aclanthology.org/2021.acl-long.201}

\bibitem{layoutlm}
Xu, Y., Li, M., Cui, L., Huang, S., Wei, F., Zhou, M.: Layoutlm: Pre-training
  of text and layout for document image understanding. In: Proceedings of the
  26th ACM SIGKDD International Conference on Knowledge Discovery \& Data
  Mining. p. 1192–1200. KDD '20, Association for Computing Machinery, New
  York, NY, USA (2020). \doi{10.1145/3394486.3403172},
  \url{https://doi.org/10.1145/3394486.3403172}

\bibitem{layoutxlm}
Xu, Y., Lv, T., Cui, L., Wang, G., Lu, Y., Florencio, D., Zhang, C., Wei, F.:
  Layoutxlm: Multimodal pre-training for multilingual visually-rich document
  understanding. arXiv preprint arXiv:2104.08836  (2021)

\bibitem{yu2020}
Yu, W., Lu, N., Qi, X., Gong, P., Xiao, R.: Pick: Processing key information
  extraction from documents using improved graph learning-convolutional
  networks  (2020), \url{https://arxiv.org/abs/2004.07464}

\end{thebibliography}
